\documentclass[runningheads]{llncs}

\usepackage{eccv}
\usepackage{eccvabbrv}
\usepackage{graphicx}
\usepackage{booktabs}
\usepackage[accsupp]{axessibility} 
\usepackage{hyperref}
\usepackage{orcidlink}
\usepackage{amsmath}
\usepackage{amssymb}
\usepackage{verbatim}
\usepackage{multirow}
\usepackage{algorithm}
\usepackage{algorithmic}
\usepackage{color}
\usepackage[dvipsnames]{xcolor}
\usepackage[capitalize]{cleveref}
\usepackage{flushend}
\usepackage{bbm}
\usepackage{bm}
\usepackage{colortbl}
\usepackage{pifont}
\newcommand{\xmark}{\ding{55}} 
\newcommand{\cmark}{\ding{51}}
\crefname{section}{Sec.}{Secs.}
\Crefname{section}{Section}{Sections}
\Crefname{table}{Table}{Tables}
\crefname{table}{Tab.}{Tabs.}

\definecolor{person}{rgb}{0.75,0.5,0.5}
\definecolor{car}{rgb}{0.5,0.5,0.5}
\definecolor{motorbike}{rgb}{0.25,0.5,0.5}
\definecolor{dog}{rgb}{0.25,0,0.5}
\definecolor{sofa}{rgb}{0,0.75,0}
\definecolor{monitor}{rgb}{0,0.25,0.5}
\definecolor{potted plant}{rgb}{0,0.25,0}
\definecolor{border}{rgb}{0.825,0.825,0.7}
\definecolor{train}{rgb}{0.5,0.75,0}
\definecolor{table}{rgb}{0.75,0.5,0}
\definecolor{sheep}{rgb}{0.5,0.25,0}
\definecolor{lightgray}{rgb}{.91,.91,.91}
\definecolor{deepred}{rgb}{0.698,0.133,0.133}
\usepackage[T1]{fontenc}
\newenvironment{myitemize}{\begin{list}{$\bullet$}
		{\setlength{\topsep}{1mm}
			\setlength{\itemsep}{0.25mm}
			\setlength{\parsep}{0.25mm}
			\setlength{\itemindent}{0mm}
			\setlength{\partopsep}{0mm}
			\setlength{\labelwidth}{15mm}
			\setlength{\leftmargin}{4mm}}}{\end{list}}

\begin{document}

\title{Cs$^2$K: Class-specific and Class-shared Knowledge Guidance for Incremental Semantic Segmentation} 


\titlerunning{Cs$^2$K: Class-specific and Class-shared Knowledge}


    \author{Wei Cong\inst{1,2,3}\orcidlink{0000-0002-9531-7179} \and
Yang Cong\inst{4, \ast}\orcidlink{0000-0002-5102-0189} \and
Yuyang Liu\inst{5}\orcidlink{0000-0002-3697-6561} \and
Gan Sun\inst{4}\orcidlink{0000-0003-1111-6909}
}

\authorrunning{W. Cong et al.}


\institute{State Key Laboratory of Robotics, Shenyang Institute of Automation, Chinese Academy of Sciences, Shenyang 110016, China \and
Institutes for Robotics and Intelligent Manufacturing, Chinese Academy of Sciences, Shenyang 110169, China \and
University of Chinese Academy of Sciences, Beijing 100049, China \and
College of Automation Science and Engineering, South China University of Technology, Guangzhou 510640, China \and
Peking University\\
\email{\{congwei45, congyang81, sungan1412\}@gmail.com, liuyuyang13@pku.edu.cn}
}

\maketitle

\footnotetext[1]{The corresponding author is Prof. Yang Cong.}

\begin{abstract}
  Incremental semantic segmentation endeavors to segment newly encountered classes while maintaining knowledge of old classes. However, existing methods either 1) lack guidance from class-specific knowledge (\emph{i.e.,} old class prototypes), leading to a bias towards new classes, or 2) constrain class-shared knowledge (\emph{i.e.,} old model weights) excessively without discrimination, resulting in a preference for old classes. In this paper, to trade off model performance, we propose the \textbf{\underline{C}}lass-\textbf{\underline{s}}pecific and \textbf{\underline{C}}lass-\textbf{\underline{s}}hared \textbf{\underline{K}}nowledge (\textbf{Cs$^2$K}) guidance for incremental semantic segmentation. Specifically, from the class-specific knowledge aspect, we design a prototype-guided pseudo labeling that exploits feature proximity from prototypes to correct pseudo labels, thereby overcoming catastrophic forgetting. Meanwhile, we develop a prototype-guided class adaptation that aligns class distribution across datasets via learning old augmented prototypes. Moreover, from the class-shared knowledge aspect, we propose a weight-guided selective consolidation to strengthen old memory while maintaining new memory by integrating old and new model weights based on weight importance relative to old classes. Experiments on public datasets demonstrate that our proposed Cs$^2$K significantly improves segmentation performance and is plug-and-play.
  
  \keywords{Incremental learning \and Semantic segmentation \and Class-specific knowledge \and Class-shared knowledge}
\end{abstract}

\section{Introduction}
\label{sec:introduction}
Semantic segmentation~\cite{semantic_segmentation_survey,robotica}, a fundamental task within the realm of computer vision~\cite{segment_any}, involves categorizing each pixel in an image to its class. Recent advancements~\cite{segformer,mask2former} have significantly enhanced the performance of semantic segmentation models, contributing to their widespread use in various applications~\cite{driving,medical}. However, adapting these models to new data streams or handling evolving classes poses challenges, as they tend to overfit new classes quickly and forget old classes during finetuning. This phenomenon is commonly known as catastrophic forgetting~\cite{catastrophic_forgetting}. Incremental semantic segmentation (ISS)~\cite{ILT, MIB, PLOP} emerges as a crucial solution to address catastrophic forgetting, focusing on maintaining knowledge about previous classes while efficiently incorporating knowledge from novel classes. We divide the stored knowledge for old classes in ISS into three forms: old class exemplars, old class features, and old model weights. Amidst the growing concerns over data privacy~\cite{incremental_survey}, we focus on the class-specific knowledge (\emph{i.e.,} old class prototypes which are the average of old class features) and the class-shared knowledge (\emph{i.e.,} old model weights) in exemplar-free methods~\cite{ILT,MIB,PLOP, SDR,RCIL,EWF}. 

\begin{figure*}[t]
  \centering
  \includegraphics[width=1\linewidth]{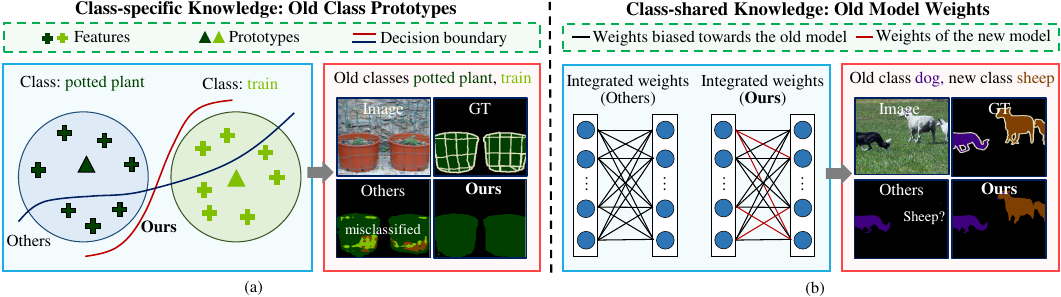} 
   \caption{Illustration of challenges for ISS. (a) The decision boundary between old classes \textcolor{potted plant}{\texttt{potted plant}} and \textcolor{train}{\texttt{train}} undergoes a dramatic change without the guidance from old class prototypes. GT means ground truth. (b) Other methods integrate old and new model weights without discrimination, leading to the integrated model weights biased towards old model weights (remember \textcolor{dog}{\texttt{dog}} but not recognize \textcolor{sheep}{\texttt{sheep}}).}
   \label{fig:motivation}
\end{figure*}

In ISS, the training dataset in one step only comprises images containing pixels belonging to the corresponding foreground classes, leading to a significantly higher proportion of these classes compared to training datasets in other steps. The discrepancies in the class distribution of different training datasets cause the overrepresentation of new classes. This phenomenon results in a dramatic change in the decision boundary, exacerbating catastrophic forgetting. As depicted in \cref{fig:motivation} (a), the segmentation model at step $t$ misclassifies the old class \textcolor{potted plant}{\texttt{potted plant}} as \textcolor{train}{\texttt{train}} and biases towards the new class \textcolor{sheep}{\texttt{sheep}}. However, most current methods~\cite{ILT,MIB,PLOP,SDR,RCIL,EWF} solely rely on class-shared knowledge (\emph{i.e.,} old model weights), which only provides limited prevention against the average forgetting of previous classes without adapting to class distribution discrepancies. In contrast, we focus on crucial class-specific knowledge (\emph{i.e.,} old class prototypes) as shown in \cref{fig:motivation} (a), which is a compact representation of the corresponding class distribution. On the one hand, current methods~\cite{PLOP,STISS,GSC} are unable to adapt to disturbances in class distribution and generate noisy pseudo labels for background (the background pixels in ISS contain the future classes, the previous classes, and the true background), thus failing to adjust the decision boundary. In this paper, we leverage the old class prototypes to correct the noisy pseudo labels. Specifically, we develop a prototype-guided pseudo labeling, which reweights the pseudo-label likelihoods assigned by the previous model, taking into account the proximity of features to prototypes. Then it corrects misclassified pixels and generates high-quality pseudo labels \textbf{from the class-specific knowledge aspect}. On the other hand, current methods ~\cite{SSUL,AMSS,alife,recall} preserving old samples to correct the decision boundary lack representative samples and leakage data privacy~\cite{incremental_survey}. To address these limitations, we design a prototype-guided class adaptation to augment old class prototypes via self-prototype augmentation and inter-prototype augmentation. Subsequently, the augmented old class prototypes are jointly trained with new classes to maintain discriminability between old and new classes \textbf{from the class-specific knowledge aspect}. The proposed prototype-guided pseudo labeling and prototype-guided class adaptation eliminate the limitations of solely relying on class-shared knowledge by fusing class-specific knowledge, thus mitigating overrepresentation of new classes.
 
Some methods~\cite{ILT,SDR,PLOP,alife} exploit regularization of class-shared knowledge (\emph{i.e.,} old model weights) to overcome catastrophic forgetting but yield limited gain since only the representations are constrained to be consistent. Other methods~\cite{RCIL,EWF} integrate the weights of the old and new model without discrimination. As illustrated in \cref{fig:motivation} (b), these approaches cause the integrated model weights biased towards the old model weights, leading to remembering the old class \textcolor{dog}{\texttt{dog}} but not recognizing the new class \textcolor{sheep}{\texttt{sheep}}. To address this issue, we introduce a weight-guided selective consolidation to simultaneously learn new classes and memorize old classes. As depicted in \cref{fig:motivation} (b), it calculates the importance of model weights for old classes based on Fisher information, then selects to integrate these important weights of the old and new models. Our weight-guided selective consolidation overcomes catastrophic forgetting while preserving new knowledge \textbf{from the class-shared knowledge aspect}.

In summary, the key contributions are:

\begin{myitemize}
\item We propose the \textbf{\underline{C}}lass-\textbf{\underline{s}}pecific and \textbf{\underline{C}}lass-\textbf{\underline{s}}hared \textbf{\underline{K}}nowledge (\textbf{Cs$^2$K}) guidance model, which is an early exploration of considering both class-specific and class-shared knowledge to surmount ISS.

\item To alleviate forgetting of old classes from the class-specific knowledge aspect, we introduce a prototype-guided pseudo labeling and a prototype-guided class adaptation to adapt to class distribution discrepancies.

\item To prevent underfitting of new classes from the class-shared knowledge aspect, we design a weight-guided selective consolidation, which selectively integrates only the crucial weights from the old model, pertaining to the old classes, into the new model to obtain obvious segmentation performance gain.
\end{myitemize}

\section{Related Works}
\subsection{Incremental Learning}
Incremental learning~\cite{icarl}, a pivotal area in machine learning, endeavors to enable models to adapt to new classes while avoiding catastrophic forgetting~\cite{catastrophic_forgetting} of previously acquired knowledge. Various strategies have been proposed in this domain, encompassing structural-based methods~\cite{DEN,PACKNET} that dynamically expand the model architecture to accommodate new classes, regularization-based methods~\cite{EWC,MAS,SI,ONLINE-EWC,LWF,spwc} employing constraints like knowledge distillation to maintain consistency of old classes, and rehearsal-based methods~\cite{icarl,GEM,FCIL,AMSS} storing or generating old samples to participate in training alongside new samples. These diverse approaches collectively aim to empower the model to incrementally acquire new knowledge while preserving previous knowledge. In this paper, we focus on challenging ISS.

\subsection{Knowledge-Guided Incremental Semantic Segmentation}
ISS~\cite{ILT,MIB,PLOP,dkd} explores to gradually adapt the segmentation model to new classes. The stored knowledge for current ISS methods mainly comprises class-specific knowledge composed of old class exemplars and old class features, as well as class-shared knowledge represented by old model weights. However, storing previous exemplars is space-intensive and privacy-insecure. Therefore, we focus on exemplar-free ISS methods. The ISS methods of storing class-specific knowledge help the model better distinguish between classes. ALIFE~\cite{alife} leverages distillation of old class features, while Incrementer~\cite{incrementer} introduces tokens for new classes. In contrast, ISS methods that store class-shared knowledge help the model overcome the average forgetting of old classes. MiB~\cite{MIB} addresses semantic drift by modeling potential classes. PLOP~\cite{PLOP} employs feature distillation with a multi-scale scheme. RCIL~\cite{RCIL} introduces average-pooling-based distillation to overcome strip pooling drawbacks. EWF~\cite{EWF} integrates the old and new model containing the old and new knowledge, respectively. Additionally, a series of methods~\cite{SSUL,microseg} introduce additional auxiliaries, making a comparison with other methods unfair. However, thoroughly combining class-specific and class-shared knowledge remains to be explored. Our method stands out by synergizing these two aspects to enhance ISS performance.


\section{Preliminaries}
\label{sec:preliminaries}
ISS sequentially learns a model $\mathcal{M}^t$ at $t\in \{0 \dots T \}$ steps, where $\mathcal{M}^t$ consisted of a feature extractor $\mathrm{\Psi}^t$ and a classifier $\mathrm{\Phi}^t$ is over parameters $\mathrm{\Theta}^t$ at step $t$. Each step $t$ involves only one dataset $\mathcal{D}^t$ consisting of input images $\bm{x}^t$ and their corresponding ground truth (GT) $\bm{y}^t$. The training GT $\bm{y}^t$ of the dataset $\mathcal{D}^t$ contains the foreground classes $C^t$ and the background $c^{bg}$. It is important to highlight that the foreground classes across steps are mutually exclusive, \emph{i.e.,} $C^i\cap C^j=\emptyset$. The ISS model continuously encountering new classes without revisiting old ones brings about the issue of catastrophic forgetting. Additionally, pixels corresponding to future, previous, and true background classes are all labeled as the background $c^{bg}$, which exacerbates catastrophic forgetting. The objective of ISS is to achieve precise segmentation for all encountered classes throughout the incremental learning process.

\begin{figure*}[t]
  \centering
  \includegraphics[width=1\linewidth]{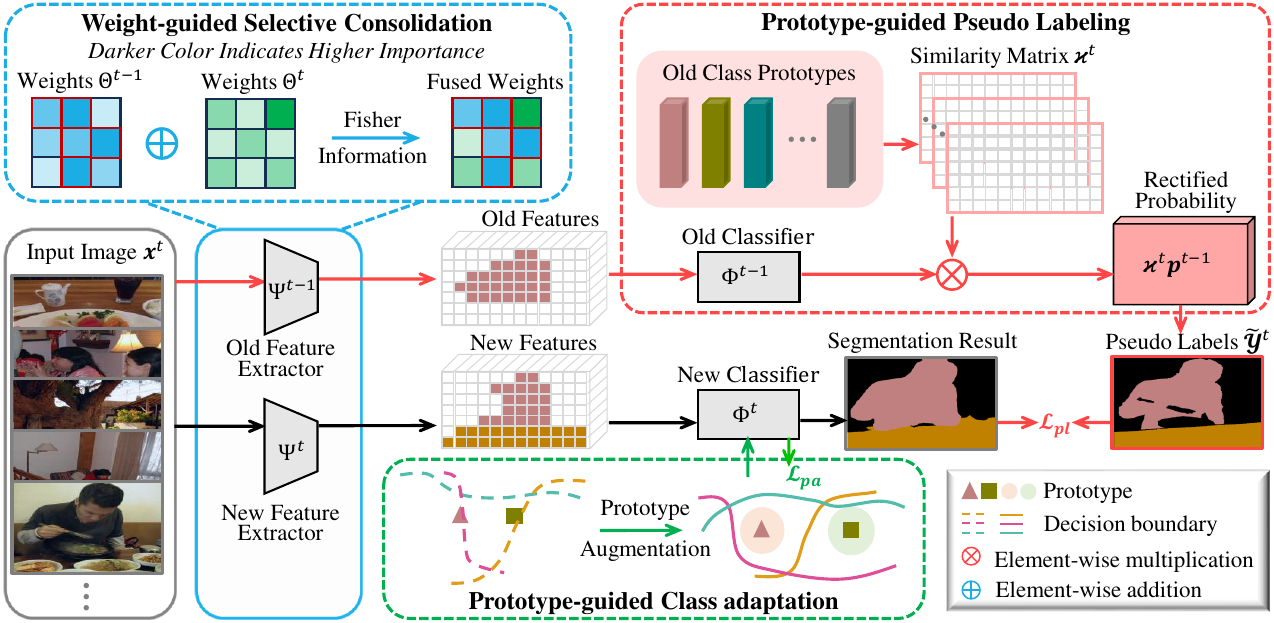} 
   \caption{Overview of our Cs$^2$K model. It updates model parameters with the proposed prototype-guided pseudo labeling and prototype-guided class adaptation from the class-specific knowledge aspect. Then, the old and new model weights are selectively integrated via the weight-guided selective consolidation to trade off performance between old and new classes from the class-shared knowledge aspect.}
   \label{fig:overview}
\end{figure*}

\section{Method}
\label{sec:method}

The overview of our Cs$^2$K model is illustrated in \cref{fig:overview}. Our Cs$^2$K model updates parameters to overcome catastrophic forgetting via the prototype-guided pseudo labeling in~\cref{sec:pgpl} and the prototype-guided class adaptation in~\cref{sec:pgca} from the class-specific knowledge aspect. Then the weight-guided selective consolidation in~\cref{sec:wgsc} is proposed to better distinguish between classes from the class-shared knowledge aspect. The overall framework is presented in \cref{sec:Overall Framework}.

\subsection{Prototype-guided Pseudo Labeling}
\label{sec:pgpl}

\begin{figure*}[t]
  \centering
  \includegraphics[width=1
\linewidth]{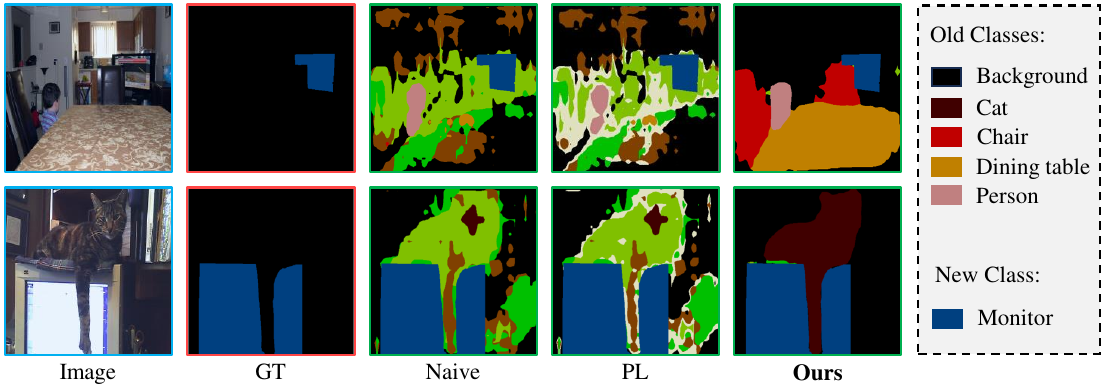} 
   \caption{The visualization comparison between pseudo labelss on Pascal VOC 2012~\cite{vocdataset}.}
   \label{fig:pseudo_label}
\end{figure*}

The discrepancies in class distribution within the training datasets of different steps cause the overrepresentation of new classes, resulting in significant changes of the decision boundary. This change leads to noisy pseudo labels~\cite{PLOP} of the background, posing a challenge since precise pseudo labels are vital for refining the decision boundary. They effectively consolidate pixels for previous classes while accommodating current class pixels. In this paper, we develop the prototype-guided pseudo labeling, as illustrated in \cref{fig:overview}, which attempts to adapt old class prototypes to correct misclassifications in pseudo labels from the class-specific knowledge aspect. We choose old class prototypes to produce high-quality pseudo labels due to the following two reasons: 1) The prototypes are not sensitive to outliers that are minority; 2) The prototypes treat classes with different occurrence frequencies equally in semantic segmentation. At step $t-1$, the calculation for the prototype $\bm{\eta}_{c}$ of the old class $c$ is formulated as:
\begin{equation}
	\bm{\eta}_{c}=\frac{\sum_{\bm{x}^{t-1} \in \mathcal{D}^{t-1}} \sum_i \mathrm{\Psi}^{t-1}(x^{t-1}_i) *\mathbbm{1}(y^{t-1}_{i,c}==1)}{\sum_{\bm{x}^{t-1} \in \mathcal{D}^{t-1}} \sum_i \mathbbm{1}(y^{t-1}_{i,c}==1)},
	\label{eq:prototype}
\end{equation}
where $x^{t-1}_i$ is the $i$-th pixel in the image $\bm{x}^{t-1}$ belonging to the dataset $\mathcal{D}^{t-1}$ at step $t-1$. $\mathrm{\Psi}^{t-1}(x^{t-1}_i)$ denotes the features of $x^{t-1}_i$ via the feature extractor $\mathrm{\Psi}^{t-1}$. $y^{t-1}_{i,c}$ represents that the GT of the pixel $x^{t-1}_i$ is the $c$-th class. $\mathbbm{1}(\cdot)$ denotes the indicator function, which gets 1 when the condition is true and 0 otherwise. The prototype $\bm{\eta}_{c}$ is the average of pixel features for the old class $c$, serving as compact representations of the corresponding class distribution. It is worth noting that we recalculate the prototype of the background at each step $t$, as its features are continuously changing. Instead of directly using old class prototypes for classification~\cite{icarl}, we try to correct pseudo labels for the previous class $c$ in background adopting the similarity weight $\kappa^t_{i,c}$ according to the old class prototype $\bm{\eta}_{c}$. Specifically, the similarity weight $\kappa^t_{i,c}$ at step $t$ that exploits feature proximity between the pixel $x^t_i$ and the old class prototype $\bm{\eta}_{c}$ is obtained as:
\begin{equation}
  \kappa^t_{i,c}=\frac{\text{exp}(-||\mathrm{\Psi}^{t-1}(x^{t}_i)-\bm{\eta}_{c}||/\tau)}{\sum_{c'\in (C^{0:t-1}\cup c^{bg})}\text{exp}\big(-||\mathrm{\Psi}^{t-1}(x^{t}_i)-\bm{\eta}_{c'}||/\tau\big)},
  \label{eq:weight_prototype}
\end{equation}
where $\tau=1$ is the temperature. $c'$ represents any previously seen old class. The similarity weight $\kappa^t_{i,c}$ reflects the confidence that the pixel $x^{t}_i$ belongs to the $c$-th class. It adapts to the disturbances of class distribution when generating pseudo labels. The formulation of the pseudo label $\tilde{y}^t_{i}$ for the pixel $x^t_i$ obtained by our proposed prototype-guided pseudo labeling is as follows:
\begin{equation}
	\tilde{y}^t_{i}=
	\left\{
	\begin{array}{lr}
		\;\;\;\;\;\;\;\;\; y^t_i\;\;\;\;\;\;\;\;\;\;\;\;\;\;\, \text{if}\; y^t_i>0 \\
         \mathop{\arg\max}\bm\kappa^t_{i}  {\bm{p}}^{t-1}_{i} \;\;\; \text{if}\; y^t_i=0 \;\text{and}\; \mathop{\arg\max}\bm\kappa^t_{i}  {\bm{p}}^{t-1}_{i}>0\\
		\;\;\;\;\;\;\;\;\;\, 0\;\;\;\;\;\;\;\;\;\;\;\;\;\;\;\, \text{otherwise},
	\end{array}
	\right.
	\label{prototype-based pseudo label}
\end{equation}
where $y^t_i$ is the GT of the pixel $x^t_i$ in the image $\bm{x}^t$. $\bm\kappa^t_{i}$ represents the similarity matrix between the pixel $x^t_i$ and old class prototypes. ${\bm{p}}^{t-1}_{i}$ denotes the softmax probability for the pixel $x^t_i$ via the model $\mathcal{M}^{t-1}$. When the GT $y^t_i >0$, it indicates that the pixel $x^t_i$ belongs to the new classes. Thus, we directly assign $y^t_i$ to the pseudo label $\tilde{y}^t_{i}$. If the GT $y^t_i$ is $0$ (\emph{i.e.,} background) and the rectified probability $\bm\kappa^t_{i}{\bm{p}}^{t-1}_{i}$ considers the pixel as an old class, the pseudo label $\tilde{y}^t_{i}$ is equal to the old class (\emph{i.e.,} $\mathop{\arg\max}\bm\kappa^t_{i}{\bm{p}}^{t-1}_{i}$). Otherwise, when the GT $y^t_i$ and the rectified probability $\bm\kappa^t_{i}{\bm{p}}^{t-1}_{i}$ both consider the pixel as $0$ (\emph{i.e.,} background), we assign $0$ to the pseudo label $\tilde{y}^t_{i}$. The prototypes have the capability to rectify misclassified pseudo labels near the decision boundary, given that the distance from pixels near the decision boundary to the corresponding prototype is much closer than to other prototypes. As shown in \cref{fig:pseudo_label}, the proposed prototype-guided pseudo labeling guides ISS model learning by generating high-quality pseudo labels compared with Naive and pseudo-labeling strategy (PL)~\cite{PLOP}. Naive means choosing the channel with the highest old probability as the pseudo label. PL~\cite{PLOP} generates pseudo labels by adopting median entropy as the measurement.

Finally, we update the ISS model using the loss $\mathcal{L}_\text{pl}$ based on the pseudo label $\tilde{\bm{y}}^t$ of $\bm{x}^t$, which is formulated as follows:
\begin{equation}
	\mathcal{L}_\text{pl}= -\frac{1}{|\mathcal{D}^t|}\sum_{\bm{x}^t \in \mathcal{D}^t}\mathcal{L}_{ce}({\bm{p}}^t, \tilde{\bm{y}}^t),\\
	\label{gradient_entropy}
\end{equation}
where $\mathcal{L}_{ce}(\cdot)$ is the function of cross entropy. ${\bm{p}}^t$ is the softmax probability of $\bm{x}^t$ by the model $\mathcal{M}^t$. $|\mathcal{D}^t|$ denotes the number of samples in the dataset $\mathcal{D}^t$.

\subsection{Prototype-guided Class Adaptation}
\label{sec:pgca}
Apart from accurate pseudo labels, replay techniques~\cite{icarl,SSUL} have been proven effective to reduce changes to the decision boundary resulting from class distribution discrepancies. Current replay strategies~\cite{SSUL,alife} in ISS select old samples that lack representativeness and leak data privacy to participate in the training. In contrast, we in this paper replay the representative old class prototypes to maintain a well-separated decision boundary from the class-specific knowledge aspect. As shown in \cref{fig:overview}, we design the prototype-guided class adaptation to optimize the shape of the decision boundary via augmented prototypes, thereby making it more adaptable to the complex distribution between different classes. Specifically, we perform prototype augmentation of the old prototype $\bm{\eta}_{c}$ in \cref{eq:prototype} through self-prototype augmentation and inter-prototype augmentation motivated by data augmentation~\cite{NAPA-VQ,Remix}. Given the considerable shift presented in background pixels, we abstain from prototype augmentation on the background. The augmented prototypes $\mathrm{\Gamma}_c$ via self-prototype augmentation is calculated by:
\begin{equation}
	\mathrm{\Gamma}_c=\bm{\eta}_{c}+\bm\mu\ast s^t,
	\label{eq:augmented_prototype}
\end{equation}
where $\bm\mu\sim\mathcal{N}(0,1)$ is the Gaussian distribution with the same dimension as the prototype $\bm{\eta}_{c}$. $s^t$ represents the scaling factor at step $t$, which is as follows:

\begin{equation}
	s^t =
	\left\{
	\begin{array}{lr}
		\;\;\;\;\;\;\;\;\;\;\;\;\;\;\;\; \sigma^{t-1}\;\;\;\;\;\;\;\;\;\;\;\;\;\;\;\;\;\;\, \text{if}\; t=1 \\
         \frac{|C^{t-1}|\ast \sigma^{t-1}+\sum\limits_{m=0}^{t-2}|C^m|\ast \sigma^{t-2}}{\sum\limits_{m=0}^{t-1}|C^m|} \;\;\; \text{if}\; t>1,
	\end{array} 
	\right.
	\label{eq:scaling_factor}
\end{equation}
where $\sigma^{t}$ is the standard deviation for the features of classes $C^t$ at step $t$. $|C^{t-1}|$ denotes the number of classes at step $t-1$. 
The scaling factor $s^t$ is a dynamic parameter that changes with step $t$, which can adaptively fit the class distribution to augment prototypes. self-prototype augmentation enhances the capability of the ISS model to thoroughly explore the feature space, mitigating the risk of being trapped in local optima. 

Then the augmented prototypes $\mathrm{\Pi}_c$ of $\bm{\eta}_{c}$ via inter-prototype augmentation is formulated as the following:
\begin{equation}
\mathrm{\Pi}_c=\lambda\ast\bm{\eta}_{c}+(1-\lambda)\ast\bm{\eta}_{c'}, \\
\emph{s.t.}\;\; c',c\in C^{0:t-1}, c'\neq c,
\label{eq:mixip}
\end{equation}
where $\lambda\sim U(0,1)$ is a random value from a uniform distribution. Performing inter-prototype augmentation can adapt to class distribution discrepancies, fostering a more balanced acquisition of distinctive features across diverse classes.

Subsequently, we update the ISS model using the loss $\mathcal{L}_{pa}$, which incorporates the augmented prototypes $\mathrm{\Gamma}_c$ and $\mathrm{\Pi}_c$ into the classifier $\mathrm{\Phi}^t$:
\begin{equation}
\begin{gathered}
\mathcal{L}_{pa}= \frac{\sum_c\Big(\mathcal{L}_{ce}\big(\mathrm{\Phi}^t(\mathrm{\Gamma}_c), y_c\big)+\lambda\ast\mathcal{L}_{ce}\big(\mathrm{\Phi}^t(\mathrm{\Pi}_c), y_c\big)+(1-\lambda)\ast\mathcal{L}_{ce}\big(\mathrm{\Phi}^t(\mathrm{\Pi}_{c}), y_{c'}\big)\Big)}{\sum\limits_{m=0}^{t-1}|C^m|} \\
 \emph{s.t.}\;\; c',c\in C^{0:t-1}, c'\neq c,
\end{gathered}
\label{eq:protoaug_loss}
\end{equation}
where $\mathrm{\Phi}^t(\mathrm{\Gamma}_c)$ and $\mathrm{\Phi}^t(\mathrm{\Pi}_c)$ represent the probabilities of the augmented prototypes $\mathrm{\Gamma}_c$ and $\mathrm{\Pi}_c$ via the classifier $\mathrm{\Phi}^t(\cdot)$ at step $t$, respectively. $y_c$ denotes the GT of the corresponding augmented prototypes $\mathrm{\Gamma}_c$ and $\mathrm{\Pi}_c$. 

\subsection{Weight-guided Selective Consolidation}
\label{sec:wgsc}
To address the challenge of class imbalance, which often favors new classes, existing methods~\cite{RCIL, EWF} tend to overly restrict the old model weights, resulting in the preference for old classes. In this paper, we propose the weight-guided selective consolidation (as depicted in \cref{fig:overview}) to selectively merge the old and new model weights based on weight importance for old classes, which effectively learns new classes while preserving the memory of previously learned classes from the class-shared knowledge aspect. In specific, the weight importance $\bm{F}^{t-1}$ is quantified by the Fisher information~\cite{gradient} of corresponding gradients at step $t-1$. After learning the step $t$, the formulation of fusing the old and new model weights based on the weight importance $\bm{F}^{t-1}$ for old classes is as follows:
\begin{equation}
\mathrm{\Theta}^t_i=\left\{
	\begin{array}{lr}
		 \omega\ast\mathrm{\Theta}^{t-1}_i+(1-\omega)\ast\mathrm{\Theta}^{t}_i\;\;\;\; \text{if}\; F^{t-1}_i>\text{TopK}(\bm{F}^{t-1}, \beta|\bm{F}^{t-1}|) \\
         \;\;\;\;\;\;\;\;\;\;\;\;\;\;\;\;\; \mathrm{\Theta}^{t}_i \;\;\;\;\;\;\;\;\;\;\;\;\;\;\;\;\; \;\;\; \text{otherwise},
	\end{array} 
	\right.
\label{eq:fisher}
\end{equation}
where $F^{t-1}_i$ represents the importance of the $i$-th old model weight for old classes. $|\bm{F}^{t-1}|$ denotes the number of weights in the model $\mathcal{M}^{t-1}$. $\text{TopK}(\bm{F}^{t-1}, \beta|\bm{F}^{t-1}|)$ represents the $\beta|\bm{F}^{t-1}|$-th largest value in $\bm{F}^{t-1}$.
$\mathrm{\Theta}^{t-1}_i$ denotes the weights of the model $\mathcal{M}^{t-1}$, containing the discriminative information for old classes. $\mathrm{\Theta}^t_i$ denotes the weights of the model $\mathcal{M}^t$, which is regarded as the best container for new classes. The selection of the number of important old model weights (\emph{i.e.,} $\beta$) and the strength of constraints applied to these weights (\emph{i.e.,} $\omega$) are crucial factors affecting the final model performance. Specifically, $\beta$ serves as the threshold to distinguish weight importance, and it is closely linked to the disparity in the quantity of classes acquired at step $t$ compared to those learned previously. Hence, the calculation of $\beta$ is designed as:
\begin{equation}
\beta=\Bigg({1+\text{exp}\bigg(\frac{|C^t|-{\sum\limits_{m=0}^{t-1}|C^m|}-1}{\sum\limits_{m=0}^{t}|C^m|+1}\bigg)}\Bigg)^{-1}. 
\label{eq:threshold_weight_importance}
\end{equation}
$\omega$ denotes the balance factor that governs the trade-off between the performance of the new and old classes, which is more associated with the ratio of the number of classes learned at step $t$ to the total number of classes encountered over time. The formulation of $\omega$ is as follows:
\begin{equation}
\omega=1-\bigg({\frac{|C^t|}{\sum\limits_{m=0}^{t}|C^m|+1}}\bigg)^{\frac{1}{2}}.  
\label{eq:omega} 
\end{equation}
$\beta$ and $\omega$ are dynamic factors, which can be automatically adjusted in various steps and scenarios. Our weight-guided selective consolidation constrains the essential model weights for old classes to overcome catastrophic forgetting, while recognizing new classes by retaining the remaining new model weights.
\subsection{Overall Framework}
\label{sec:Overall Framework}
The ISS model is updated continually with the proposed prototype-guided pseudo labeling and the prototype-guided class adaptation from the class-specific knowledge aspect when learning new classes. The overall loss $\mathcal{L}$ is formulated as follows:
\begin{equation}
	\mathcal{L}= \mathcal{L}_{pl}+\mathcal{L}_{pa}.\\
	\label{s2kg}
\end{equation}
After learning new classes, the old and new model weights are selectively integrated via the weight-guided selective consolidation from the class-shared knowledge aspect. Combining the above techniques, our Cs$^2$K model effectively learns new classes without forgetting previously learned ones.

\section{Experiments}
\label{sec:experiments}

\begin{table*}[t]
\caption{Comparison on Pascal VOC 2012~\cite{vocdataset}. \textbf{\color{purple}Red} highlights the highest results.}
\setlength\tabcolsep{8pt}
\centering
\small
\scalebox{0.815}{
\begin{tabular}{ccc|c|cc|c|cc|c}
\toprule
\multirow{2}{*}{\textbf{Method}}     & \multicolumn{3}{|c|}{\textbf{15-1 (6 steps)}} & \multicolumn{3}{c|}{\textbf{10-1 (11 steps)}}                           & \multicolumn{3}{c}{\textbf{5-3 (6 steps)}} \\ 
  & \multicolumn{1}{|c} {0-15}        & 16-20         & all        & 0-10         & 11-20          & all         & 0-5       & 6-20       & all   \\ 
\midrule
\multicolumn{1}{c|}{FT~\cite{deeplab}}             & \, 0.2        & \, 1.8        & \, 0.6       & \, 6.3         & \, 1.1         & \, 3.8       & 11.8       & \, 5.2        & \, 7.1     \\
\multicolumn{1}{c|}{Joint~\cite{deeplab}}           & 79.5        & 74.0       & 78.2       & 79.0         & 77.3        & 78.2       & 78.0         & 78.3        & 78.2        \\

\multicolumn{1}{c|}{LWF~\cite{LWF}}             & \, 6.0        & \, 3.9        & \, 5.5       & \, 8.0         & \, 2.0         & \, 4.8       & 20.9       & 36.7        & 24.7   \\ 
\multicolumn{1}{c|}{ILT~\cite{ILT}}             & \, 9.6        & \, 7.8       & \, 9.2       & \, 7.2         & \, 3.7        & \, 5.5       & 22.5       & 31.7        & 29.0  \\

\multicolumn{1}{c|}{SDR~\cite{SDR}}             & 47.3        & 14.7       & 39.5       & 32.4         & 17.1        & 25.1       & -       & -        & -    \\ 
\multicolumn{1}{c|}{RCIL~\cite{RCIL}}        & 70.6        & 23.7       & 59.4       & 55.4         & 15.1        & 34.3       & 63.1       & 34.6        & 42.8         \\ 
\multicolumn{1}{c|}{GSC~\cite{GSC}}        & 72.1        & 24.4       & 60.8       & 50.6         & 17.3        & 34.7       & 32.7       & 30.1        & 30.9      \\ \midrule
\multicolumn{1}{c|}{MiB~\cite{MIB}}             & 38.0        & 13.5       & 32.2       & 12.2         & 13.1        & 12.6       & 57.1       & 42.5        & 46.7     \\ 
\multicolumn{1}{c|}{MiB+EWF~\cite{EWF}}        & 78.0        & 25.5       & 65.5      & 56.0         & 16.7        & 37.3       & 69.0      & 45.0        & 51.8   \\
\rowcolor{lightgray} \multicolumn{1}{c|}{MiB+Cs$^2$K (\textbf{Ours})}   & 76.2        & 41.8       & 68.0      & 43.0         & 35.2        & 39.3       & 70.6           & 50.4            & \textbf{\color{purple}56.2}    \\ \midrule
\multicolumn{1}{c|}{PLOP~\cite{PLOP}}            & 65.1        & 21.1       & 54.6       & 44.0         & 15.5        & 30.5       & 25.7       & 30.0        & 28.7  \\ 
\multicolumn{1}{c|}{PLOP+EWF~\cite{EWF}}        & 77.7        & 32.7       & 67.0      & 71.5         & 30.3        & 51.9       & 61.7       & 42.2        & 47.7   \\
\rowcolor{lightgray} \multicolumn{1}{c|}{PLOP+Cs$^2$K (\textbf{Ours})}   & 77.9        & 46.4       & \textbf{\color{purple}70.4}       & 74.4         & 47.2       & \textbf{\color{purple}61.5}       & 58.4           & 53.4            & 54.8     \\ 

\bottomrule

\end{tabular}
}
\label{tab:voc_base}
\end{table*}

\subsection{Experimental Setups} 
\label{sec:exp_set}

\textbf{Evaluation Protocols.} The ISS training process is typically divided into $T$ steps, with each step representing an individual task, and the labeled classes within each step are disjoint. We adhere to  the widely-used \textit{overlapped} setting, as adopted in previous works~\cite{EWF,star}. This choice stems from the acknowledgment that within the current training, the background class contains both old and future classes. Following previous methods~\cite{MIB,PLOP,EWF}, we perform experiments on two public datasets: PASCAL VOC 2012~\cite{vocdataset} and ADE20K~\cite{adedataset}. The former comprises 20 distinct classes and the background class, while the latter consists of 150 classes. We evaluate the effectiveness under 15-1, 10-1, and 5-3 scenarios on Pascal VOC 2012~\cite{vocdataset}. Additionally, we conduct experiments on ADE20K~\cite{adedataset} under 100-10 and 100-5 scenarios. The X-Y scenario indicates learning X classes in the first step, followed by learning Y classes in the subsequent steps. At each step, we only access the current data. Moreover, we adopt mIoU as the metric.

\noindent \textbf{Implementation Details.} Following popular works~\cite{MIB,PLOP,EWF}, our architecture utilizes Deeplab-v3~\cite{deeplab} with a ResNet-101~\cite{resnet} pre-trained on ImageNet~\cite{imagenet}. We ensure that details such as learning rate, batch size, optimizer, and dataset processing remain consistent with previous methods~\cite{PLOP,EWF}. Due to the commonality in the first step across all methods, we reuse the weights acquired during this phase. We conduct experiments on four NVIDIA RTX 3090 GPUs.

\begin{figure*}[t]
  \centering
  \includegraphics[width=1
\linewidth]{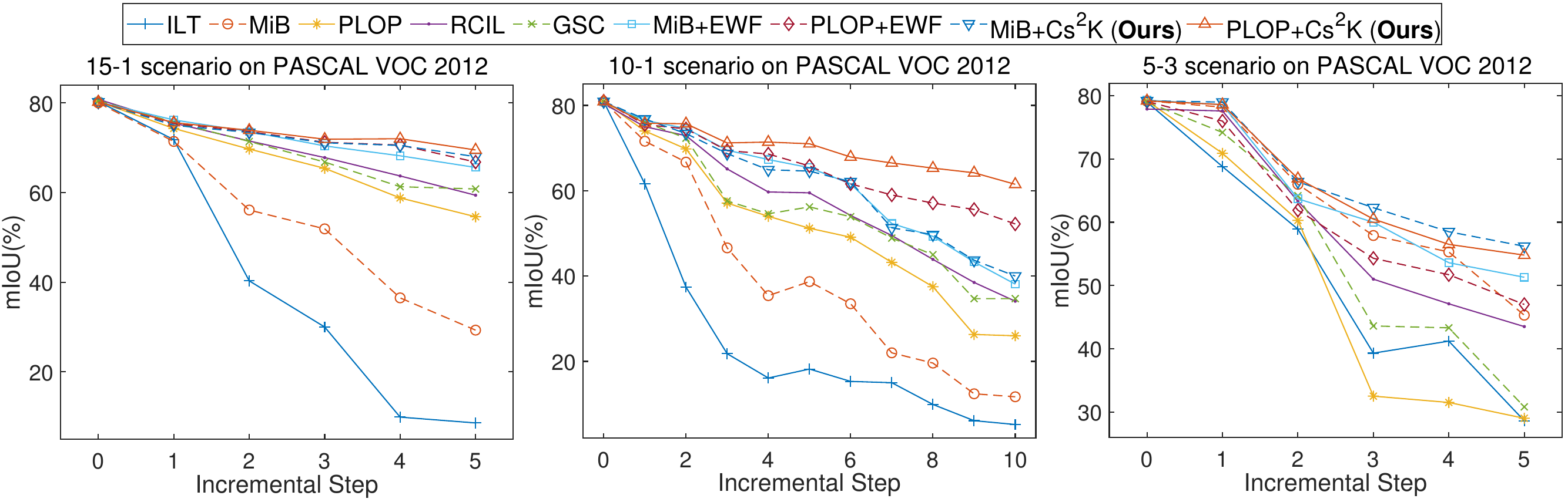} 
   \caption{Quantitative comparison at each step with different methods for 15-1, 10-1, and 5-3 class incremental segmentation scenarios on Pascal VOC 2012~\cite{vocdataset}.}
   \label{fig:he}
\end{figure*}

\begin{figure*}[ht]
  \centering
  \includegraphics[width=1
\linewidth]{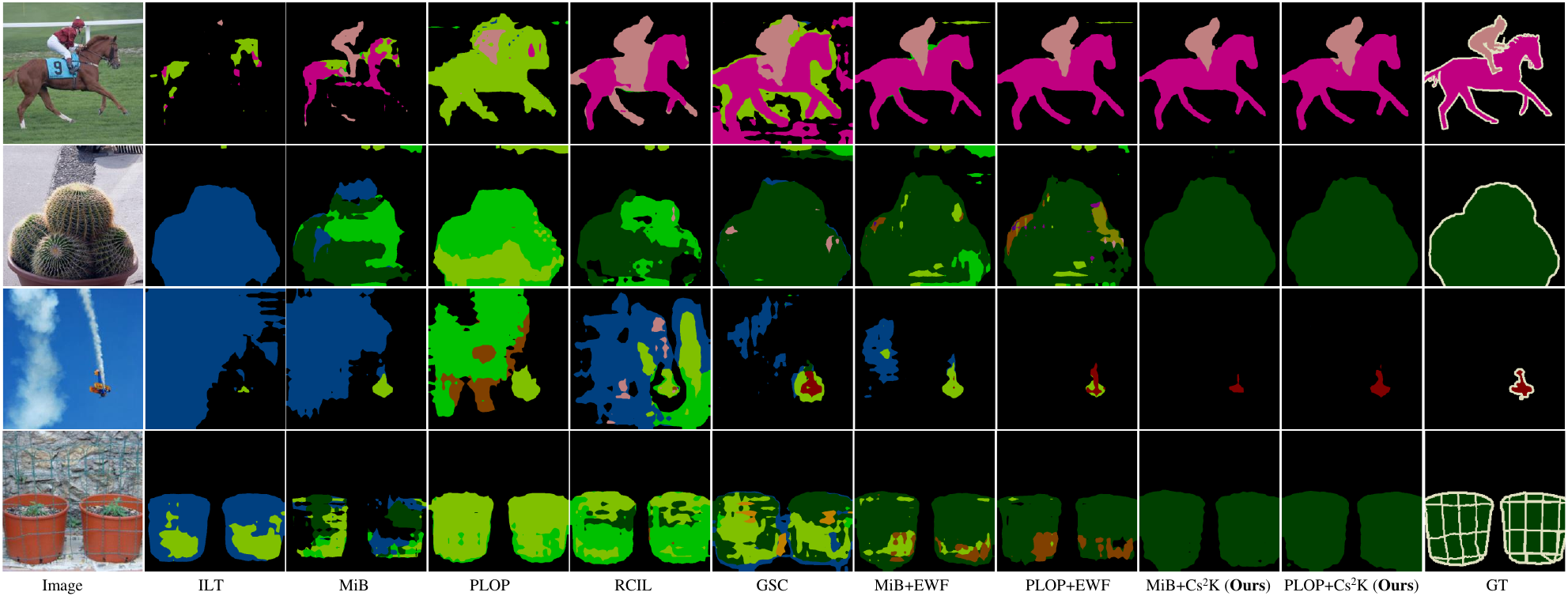} 
   \caption{The visualization comparison from the last step on Pascal VOC 2012~\cite{vocdataset}.}
   \label{fig:vis}
\end{figure*}

\subsection{Comparisons}
\label{sec:comparisons}
In our comparative analysis, we benchmark our Cs$^2$K model with the classic continual learning method LWF~\cite{LWF} and several ISS algorithms ILT~\cite{ILT}, MiB~\cite{MIB}, PLOP~\cite{PLOP}, SDR~\cite{SDR}, RCIL~\cite{RCIL}, GSC~\cite{GSC}, and EWF~\cite{EWF}. Notably, EWF~\cite{EWF} is applied to MiB~\cite{MIB} and PLOP~\cite{PLOP} following the original paper. FT~\cite{deeplab} serves as a lower bound, training only on the newly encountered data. Joint~\cite{deeplab} trains on all seen classes, which is an upper bound. Additionally, our proposed Cs$^2$K is plug-and-play and we apply it to MiB~\cite{MIB} and PLOP~\cite{PLOP} for evaluation.

\begin{figure*}[t]
  \centering
  \includegraphics[width=1
\linewidth]{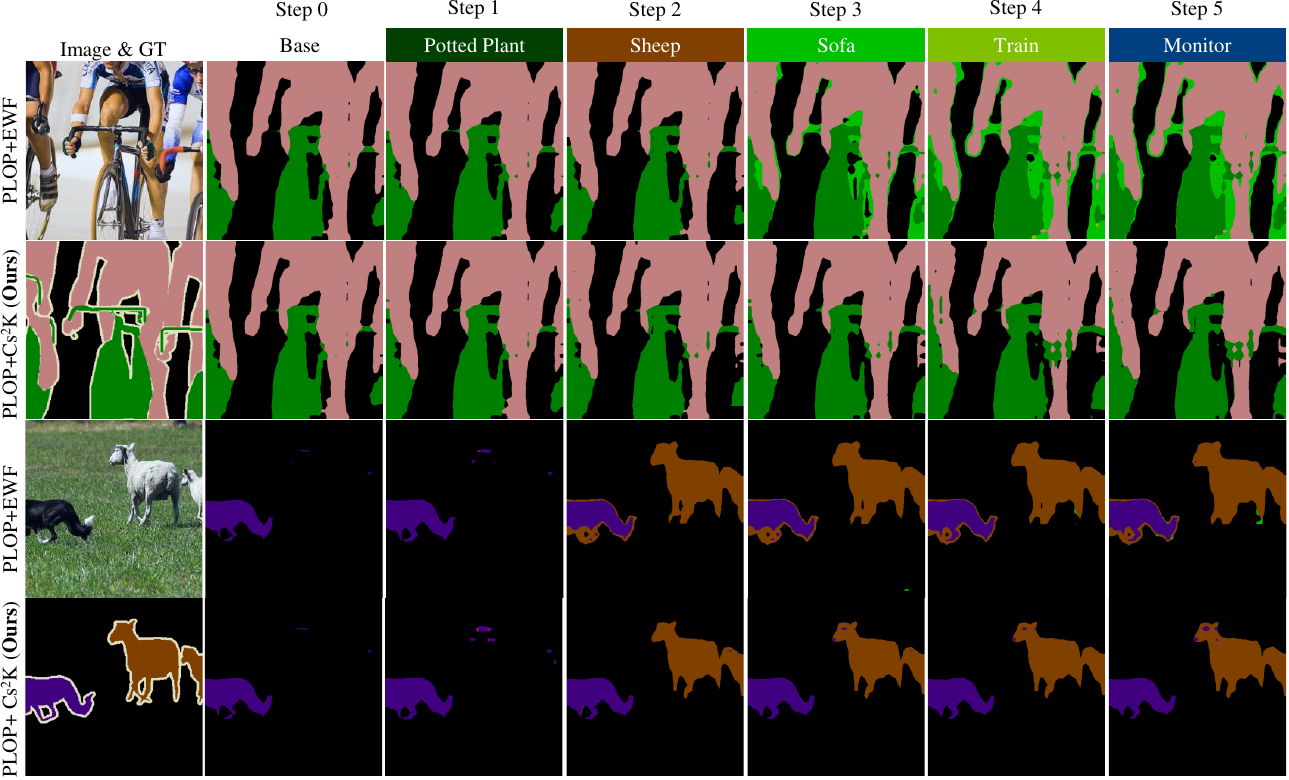} 
   \caption{The visualization comparison across steps on Pascal VOC 2012~\cite{vocdataset}.}
   \label{fig:each_step}
\end{figure*}

\begin{table*}[ht]
\caption{Comparison on ADE20K~\cite{adedataset}. \textbf{\color{purple}Red} highlights the highest results.}
\setlength\tabcolsep{3.2pt}
\centering
\small
\scalebox{0.815}{
\begin{tabular}{ccccccc|c|cc|c}
\toprule
  \multirow{2}{*}{\textbf{Method}}     & \multicolumn{7}{|c|}{\textbf{100-10 (6 steps)}}                              & \multicolumn{3}{c}{\textbf{100-5 (11 steps)}} \\ 
   &\multicolumn{1}{|c} {1-100} & 101-110 & 111-120 & 121-130 & 131-140 & 141-150 & all   & 1-100      & 101-150       & all   \\ \midrule
\multicolumn{1}{c|}{Joint~\cite{deeplab}}                      & 44.3       & 26.1        & 42.8     & 26.7  & 28.1    & 17.3    & 38.9 & 44.3   & 28.2    & 38.9    \\ 
\multicolumn{1}{c|}{ILT~\cite{ILT}}                      &  \, 0.1   & \, 0.0     & \, 0.1     & \, 0.9     & \, 4.1     & \, 9.3     & \, 1.1  & \, 0.1   & \, 1.3    & \, 0.5       \\ 

\multicolumn{1}{c|}{PLOP~\cite{PLOP}}                   & 40.6  & 15.2    & 16.9    & 18.7    & 11.9   & \, 7.9     & 31.6 & 39.1   & \, 7.8  & 28.7     \\ 
\multicolumn{1}{c|}{RCIL~\cite{RCIL}}                & 39.3  & 14.6    & 26.3    & 23.2    & 12.1    & 11.8    & 32.1  & 38.5   & 11.5    & 29.6    \\
\multicolumn{1}{c|}{GSC~\cite{GSC}}                & 40.8  & 14.3    & 24.6    & 22.2    & 15.2    & 11.7    & 32.6 & 39.5   & 11.2    & 30.2     \\ \midrule
\multicolumn{1}{c|}{MiB~\cite{MIB}}               & 38.3  & 12.6    & 10.6    & \, 8.7     & \, 9.5     & 15.1    & 29.2 & 36.0   & \, 5.6    & 25.9    \\ 
\multicolumn{1}{c|}{MiB+EWF~\cite{EWF}}               & 41.5  & 12.8    & 22.5    & 23.2    & 14.4    & \, 8.8    & 33.2 & 41.4   & 13.4    & 32.1     \\ 
\rowcolor{lightgray}\multicolumn{1}{c|}{MiB+Cs$^2$K (\textbf{Ours})}               & 42.4  & 11.6    & 29.4    & 22.8    & 14.5    & \, 7.7    & \textbf{\color{purple}34.1} & 41.9   & 18.4    & \textbf{\color{purple}34.2}      \\

\bottomrule
\end{tabular}}
\label{tab:ade2}
\end{table*}

\noindent \textbf{Pascal VOC 2012.} \cref{tab:voc_base} presents the comparison results of challenging 15-1, 10-1 and 5-3 scenarios on PASCAL VOC 2012~\cite{vocdataset}. We observe that our method surpasses MiB~\cite{MIB} and PLOP~\cite{PLOP} by a substantial margin in all scenarios, achieving notable mIoU gains of 35.8\% and 31.0\% respectively in the 15-1 and 10-1 scenarios. Besides, comparing with advanced methods still significantly highlights our advantages. Specifically, our method surpasses the latest advancements PLOP+EWF~\cite{EWF} and MiB+EWF~\cite{EWF} by 9.6\% and 4.4\% mIoU in the 10-1 and 5-3 scenarios, respectively. This underscores the effectiveness of correcting decision boundaries for better distinguishing between classes from the class-specific knowledge aspect. Furthermore, our method maintains comparable performance on previous classes and achieves excellent performance on current classes. Compared to PLOP+EWF~\cite{EWF}, we observe substantial mIoU improvements of 13.7\%, 16.9\%, and 11.2\% on new classes in the 15-1, 10-1, and 5-3 scenarios, respectively. It is attributed to the discriminative integration of weights between the old and new model, effectively preserving new knowledge from the class-shared knowledge aspect. The comparison at each step for 15-1, 10-1, and 5-3 scenarios is shown in \cref{fig:he}. Our method consistently maintains a leading performance, especially in the challenging 10-1 scenario. This illustrates our robustness in combining the class-specific and class-shared knowledge.

\begin{table*}[t]
\caption{Ablation study of different pseudo label strategies on PASCAL VOC 2012~\cite{vocdataset}. \textbf{\color{purple}Red} highlights the highest results.}
\setlength\tabcolsep{5.5pt}
\centering
\small
\scalebox{0.815}{
\begin{tabular}{c|ccccccccccc}
\toprule
\multirow{2}{*}{\textbf{Pseudo label Strategy}}     & \multicolumn{11}{c}{\textbf{Step}}\\
& 0 & 1 & 2 & 3 & 4 & 5 & 6 & 7 &  8 &  9 &  10 \\
\midrule
Naive             & 80.9       & 75.8       & 75.5       &  71.2      & 71.4   & 69.9     & 67.8       & 65.7       &  63.9      & 61.5   & 58.3       \\ 
PL~\cite{PLOP}             & 80.9       &  75.5      &  75.5      &  71.1      & 71.3   & 71.0  & 66.3       &  60.9      &  57.9      &  57.2  & 54.0 \\  \midrule
\rowcolor{lightgray}Cs$^2$K (\textbf{Ours})            &  80.9      & 75.8       & 75.7     & 71.2      & 71.4   & 71.0    & 67.9       & 66.5       & 65.3       & 64.2   & \textbf{\color{purple}61.5}         \\ \bottomrule
\end{tabular}}
\label{tab:PL strategy}
\end{table*}

\begin{table*}[ht]
\caption{Ablation study of weight integration strategies on PASCAL VOC 2012~\cite{vocdataset}. \textbf{\color{purple}Red} highlights the highest results.}
\setlength\tabcolsep{7pt}
\centering
\small
\scalebox{0.81}{
\begin{tabular}{c|ccccccccccc}
\toprule
\multirow{2}{*}{\textbf{Fusion Strategy}}     & \multicolumn{11}{c}{\textbf{Step}}\\
 &0 & 1 & 2 & 3 & 4 & 5 & 6 & 7 &  8 &  9 &  10 \\ \midrule
WF~\cite{EWF}             &  80.9      &  75.7      & 75.6       & 71.1       &  71.0  & 69.7     & 67.6       &  66.1      &  64.7      & 63.0   & 60.3        \\ \midrule
\rowcolor{lightgray}Cs$^2$K (\textbf{Ours})            &  80.9      & 75.8       & 75.7     & 71.2      & 71.4   & 71.0    & 67.9       & 66.5       & 65.3       & 64.2   & \textbf{\color{purple}61.5}        \\ \bottomrule
\end{tabular}}
\label{tab:fusion strategy}
\end{table*}

\noindent \textbf{Visualization.} \cref{fig:vis} visually showcases the results of the final step in the 15-1 scenario. Unlike previous methods that display varying degrees of misclassification for old classes, our method excels in producing accurate segmentation. This intuitively indicates that our proposed method consistently exhibits outstanding performance. Additionally, ~\cref{fig:each_step} presents the visualization results across steps in the 15-1 scenario. Both methods generate the same visualization results as there is no distinction at the first step. However, MiB+EWF~\cite{EWF} rapidly forgets previous classes, showing a bias towards new classes. In contrast, our Cs$^2$K exhibits greater stability. It is attributed to our prototype-guided pseudo labeling, prototype-guided class adaptation, and weight-guided selective consolidation.

\noindent \textbf{ADE20K.} We further validate our method on ADE20K~\cite{adedataset}. Experiments are conducted on the most challenging scenarios, 100-10 and 100-5, while discarding the less meaningful 100-50 scenario. Our method, as depicted in \cref{tab:ade2}, consistently surpasses all other competing methods across all scenarios on ADE20K~\cite{adedataset}. For instance, our method attains a 2.1\% mIoU increase compared to MiB+EWF~\cite{EWF} in the 100-5 scenario. This underscores the robustness and generalizability of our proposed method on the more realistic dataset.

\subsection{Ablation Study}
\label{sec:ablation study}

\noindent \textbf{Pseudo Label Strategy.} To validate the effectiveness of our prototype-guided pseudo labeling, we compare it with Naive and the pseudo-labeling strategy (PL)~\cite{PLOP}. Naive determines the pseudo label by adopting the channel with the highest old probability. \cref{fig:pseudo_label} presents the visualization results. Naive generates the noisiest pseudo labels, while PL removes some uncertain pixels based on the median entropy. 
When the model exhibits strong performance, removing uncertain pseudo labels leads to poor results. In contrast, our method, guided by the correction of representative prototypes, consistently produces accurate pseudo labels. Quantitative results in \cref{tab:PL strategy} align with the visualization.

\noindent \textbf{Weight Integration Strategy.} 
We compare our weight-guided selective consolidation with the weight fusion (WF)~\cite{EWF} which equally constrains all old model weights to overcome catastrophic forgetting. However, as shown in \cref{tab:fusion strategy}, WF~\cite{EWF} yields suboptimal results. In contrast, our method attains superior performance, underscoring the effectiveness of selectively integrating crucial previous model weights.

\begin{table}[htbp]
\caption{Ablation study of the 15-1 scenario. \textbf{\color{purple}Red} highlights the highest results.}
\setlength\tabcolsep{10pt}
\centering
\small
\scalebox{0.84}{
\begin{tabular}{c|cccc|cc|c}
\toprule
\multirow{2}{*}{Settings} &\multicolumn{4}{c|}{Variants} & \multicolumn{3}{c}{15-1 Scenario}  \\ 
      & PPL      & PCA-SA      & PCA-IA  &WSC  & 0-15      & 16-20      & all       \\
\bottomrule
Ours-w/o PPL        & \xmark        & \cmark  & \cmark  &\cmark  & 70.4          & 49.0           &  65.3             \\ 
Ours-w/o PCA        & \cmark       & \xmark   & \xmark  &\cmark  & 78.5          & 37.5           & 68.7 
\\
Ours-w/o PCA-SA        & \cmark       & \xmark   & \cmark  &\cmark &   77.3        &  42.2         &  69.0
\\ 
Ours-w/o PCA-IA      & \cmark       & \cmark  & \xmark  &\cmark   &  77.8         & 43.7           & 69.7
\\
Ours-w/o WSC        & \cmark       & \cmark   & \cmark  & \xmark   &   58.0        & 18.8           &  48.6         \\ 
\rowcolor{lightgray}Cs$^2$K (\textbf{Ours})        & \cmark       & \cmark   & \cmark &\cmark  &  77.9         &  46.4          & \textbf{\color{purple}70.4}              \\ \bottomrule
\end{tabular}}
\label{tab:ablation study}
\end{table}

\noindent \textbf{Ablation Study on Proposed Components of Cs$^2$K.} To further assess the impact of the introduced prototype-guided pseudo labeling (PPL), prototype-guided class adaptation (PCA), and weight-guided selective consolidation (WSC), we perform the ablation study in the 15-1 scenario on PASCAL VOC 2012~\cite{vocdataset}. We refer to PCA with only self-prototype augmentation as PCA-SA, and PCA with only inter-prototype augmentation as PCA-IA. \cref{tab:ablation study} reveals that our proposed PPL can improve performance by 5.1\% mIoU, indicating that prototype correction can generate high-quality pseudo labels. Our PCA achieves 1.7\% mIoU gain with effective prototype augmentation. This illustrates its ability to adapt to class distribution discrepancies and distinguish between old and new classes. Additionally, our proposed WSC improves the performance by 21.8\% mIoU, highlighting the necessity of selectively consolidating important old knowledge for ISS. In summary, each component in our Cs$^2$K has been proven effective, and their simultaneous utilization contributes to the overall superior performance.

\section{Conclusion}
In this paper, we mitigate catastrophic forgetting of old classes and underfitting of new classes caused by neglecting the class-specific knowledge and equally treating the class-shared knowledge. We propose the \textbf{\underline{C}}lass-\textbf{\underline{s}}pecific and \textbf{\underline{C}}lass-\textbf{\underline{s}}hared \textbf{\underline{K}}nowledge (\textbf{Cs$^2$K}) guidance to surmount ISS. The prototype-guided pseudo labeling and prototype-guided class adaptation are designed to adapt to class distribution discrepancies from the class-specific knowledge aspect. Then the weight-guided selective consolidation is proposed to distinguish between classes from the class-shared knowledge aspect. 
Our effectiveness is rigorously validated through extensive experiments on public benchmark datasets. While acknowledging a performance gap compared to joint training in long sequence tasks, we emphasize that this serves as a foundation for future endeavors, where we aim to further investigate and bridge this gap.

\section*{Acknowledgments}
This work is supported by National Key R\&D Program of China (2023YFB4704800), and National Nature Science Foundation of China under Grant (62225310, 62127807).

%
%
\bibliographystyle{splncs04}

\end{document}